\documentclass[letterpaper, 10 pt, conference]{ieeeconf}  % Comment this line out if you need a4paper
\usepackage{graphicx}
\usepackage[hidelinks]{hyperref}
\usepackage{amsmath}
\usepackage{amssymb}
\usepackage{subcaption}
\usepackage{cite}
\usepackage{booktabs}  % For better table formatting
\usepackage{siunitx}   % For scientific notation
\usepackage[font=small]{caption}
\usepackage{comment}
\usepackage{url}
\IEEEoverridecommandlockouts                              % This command is only needed if 
\usepackage{xcolor}
\usepackage{soul} % for \st and \setstcolor used in comments

\xdefinecolor{redZ}{RGB}{234,29,93}
\xdefinecolor{blueZ}{RGB}{19,106,213}
\xdefinecolor{yellowZ}{RGB}{251,138,46}
\xdefinecolor{greenYL}{RGB}{0, 191, 0}

\title{\LARGE \bf
    RCM Constraint-Consistent Dynamic Control in Surgical Robots
}

 \author{Yu Li$^{1}$, Hamid Sadeghian$^{1}$, Zewen Yang$^{1}$, Valentin Le Mesle$^{1}$, and Sami Haddadin$^{2}$ 
  \thanks{$^{1}$ Munich Institute of Robotics and Machine Intelligence, Technical University of Munich, Germany. Corresponding Author's e-mail: yu.li@tum.de}\thanks{$^{2}$ Mohamed Bin Zayed University of Artificial Intelligence, Abu Dhabi, UAE.}
  \thanks{
The authors would like to thank the Federal Ministry of Research, Technology, and Space (BMFTR) for its support as part of the research program Communication Systems ``Souver\"an. Digital. Vernetzt.''. Joint project 6G-life, project identification number: 16KIS2414.
This work has also received funding from the European Union's Horizon Europe research and innovation programme as part of the project FlexCycle under grant agreement No. 101189600.
  }
}

\begin{document}

\maketitle
\thispagestyle{empty}
\pagestyle{empty}

%%%%%%%%%%%%%%%%%%%%%%%%%%%%%%%%%%%%%%%%%%%%%%%%%%%%%%%%%%%%%%%%%%%%%%%%%%%%%%%%
% \begin{abstract}
% Robotic-assisted minimally invasive surgery imposes strict motion constraints dictated by the Remote Center of Motion (RCM), requiring specialized control strategies. 
% While existing methods assume RCM as pure control objectives, they often lack interpretation of kinematic constraints and fail to account for rheonomic variations in trocar positioning. 
% This work introduces a constraint-projected inverse dynamics control formulation that ensures precise task execution and satisfies the RCM constraint.
% The proposed control mitigates task-based and kinematic constraints into projection-based orthogonal decomposed control, enabling improved energy efficiency and robustness against disturbances. 
% Experimental results validate improved tracking accuracy and lower torque consumption compared to classical null-space projection methods while maintaining effective null-space compliance during human interaction.
% \end{abstract}

\begin{abstract}
Robotic-assisted minimally invasive surgery (RAMIS) requires accurate enforcement of the remote center of motion (RCM) constraint to ensure safe tool motion through a trocar. Existing virtual RCM controllers are commonly formulated either at the kinematic level or as task-space objectives, which makes torque-level enforcement under trocar motion and physical interaction difficult to formulate consistently. This paper models the RCM as a rheonomic holonomic constraint and incorporates it into a projection-based inverse-dynamics controller with explicit constrained/free-motion torque decomposition. The resulting formulation unifies kinematic RCM enforcement and task-space tracking at the torque level, while preserving a constraint-consistent structure for residual regulation and null-space compliance. The proposed controller is validated in simulation and on a RAMIS training platform against representative projection-based and constrained-dynamics baselines. Across spiral tracking, varying insertion depth, moving trocar conditions, and human interaction, the method achieves lower RCM residuals and smoother torque profiles while maintaining accurate tool-tip tracking. These results support the use of constraint-consistent torque control for reliable virtual RCM enforcement in surgical robotics. The project page is available at \url{https://rcmpc-cube.github.io}.
\end{abstract}

%%%%%%%%%%%%%%%%%%%%%%%%%%%%%%%%%%%%%%%%%%%%%%%%%%%%%%%%%%%%%%%%%%%%%%%%%%%%%%%%

\section{Introduction}
Minimally invasive surgery (MIS) reduces trauma, recovery time, and infection risk by operating through small incisions~\cite{smith2005robotic}. Despite these benefits, MIS remains demanding due to constrained access and safety-critical motion~\cite{nugent2012evaluation}. Robotic-assisted MIS (RAMIS) enhances dexterity and precision while reducing surgeon workload through motion scaling and tremor filtering~\cite{smith2005robotic}, and further supports training and assessment via teleoperation and haptic feedback~\cite{elek2019robot}. Central to RAMIS safety is the remote center of motion (RCM), which constrains tool motion about a trocar pivot~\cite{funda1996constrained}. Enforcing this constraint with high precision under dynamic and interactive conditions remains a major control challenge.

Early solutions relied on mechanical RCM mechanisms, e.g., parallelogram linkages, spherical joints, and hybrid architectures~\cite{aksungur2015remote,wang2022family}, which reliably maintain pivoting but limit adaptability and workspace flexibility. This has motivated control-based enforcement. Among these, kinematic methods incorporate constrained Jacobians~\cite{sadeghian2019constrained}, parameterize tool motion~\cite{aghakhani2013task}, or solve inverse kinematics and velocity control~\cite{davila2024real}. Data-driven variants employ recurrent models for servoing and parameter estimation~\cite{li2020accelerated,liu2024data}. However, these methods are often tailored to position or velocity control and remain less compatible with torque-level implementations, where dynamic consistency and robustness are essential~\cite{minelli2021dynamic}.

Implicit formulations mitigate some of these limitations. Projection-based and null-space schemes~\cite{kastritsi2021control} enforce virtual RCM constraints implicitly by restricting task velocities, while multi-priority control extends this idea to dynamical task constraints via augmented Jacobians~\cite{sandoval2017new}. In collaborative RAMIS~\cite{su2019improved}, dynamic controllers have further incorporated disturbance compensation~\cite{su2022fuzzy} as well as optimal and passive teleoperation schemes~\cite{piccinelli2024passive,kastritsi2024passive}. More recently, constrained-dynamics formulations~\cite{udwadia2002foundations,minelli2022torque} have explicitly separated constrained and free dynamics, providing a more consistent dynamic treatment. However, previous virtual RCM controllers remain primarily formulated at the kinematic or task level, and their extension to rheonomic or enforced RCM conditions induced by trocar motion and viscoelastic tissue behavior is less explicit~\cite{kirilova2009visco}.

\begin{figure}[!t]
    \centering
    \vspace{3mm}
    \includegraphics[width=0.8\linewidth]{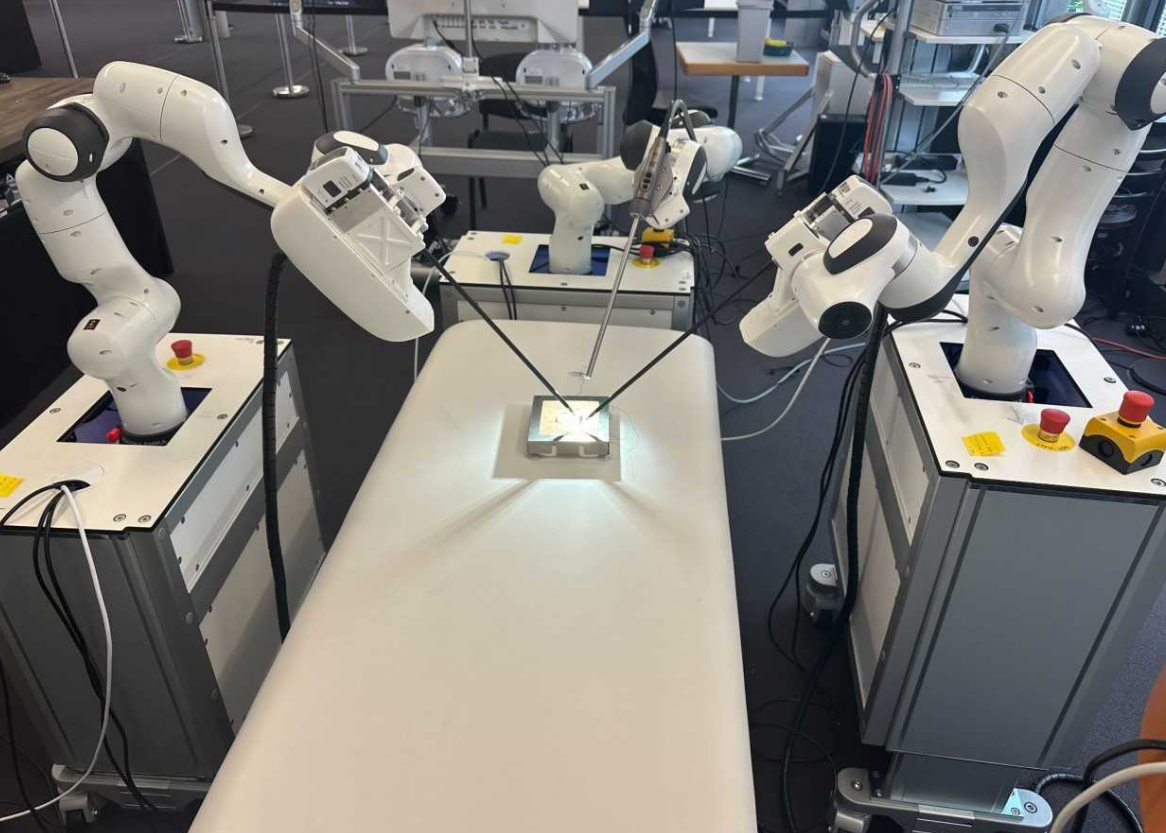}
    \caption{RAMIS training platform used in this work. The setup comprises three surgical robot manipulators (SRMs): one for endoscope positioning and two equipped with the surgical tool. The platform can also be operated in teleoperation mode.} 
    \label{fig:surgical-platform}
    \vspace{-5mm}
\end{figure}

More broadly, projected constrained dynamics and operational-space control have proven effective in contact-rich robotics~\cite{khatib1987unified,mistry2012operational}. Orthogonal projections can reduce torque effort~\cite{righetti2011inverse} and accommodate friction-aware constraints through optimization-based formulations~\cite{aghili2016control,wang2023towards}. However, classical projection-based methods typically assume rigid scleronomic constraints and remain sensitive to modeling errors and Jacobian conditioning~\cite{aghili2005unified,moura2019equivalence}, which limits their direct applicability to RCM control under trocar motion and non-ideal abdominal-wall compliance. Relative to augmented-Jacobian virtual RCM control~\cite{sandoval2017new} and constrained-dynamics decomposition~\cite{minelli2022torque}, the present formulation explicitly embeds rheonomic virtual RCM regulation into a projection-based constraint-consistent torque-control framework.

%This work reinterprets RCM as a rheonomic holonomic constraint and embeds it into a projection-based inverse-dynamics framework, reconciling task-level and kinematic formulations in a constraint-consistent manner~\cite{moura2019equivalence}. This unification enables accurate tool-tip tracking, smoother and more efficient torque behavior, and reliable RCM constraint satisfaction. % under clinically relevant scenarios. We validate the proposed controller in simulation and on a RAMIS training platform, benchmarking against state-of-the-art methods under spiral trajectories, variable insertion depths, trocar motion, and human interaction.
The main contributions of this paper are threefold:

\begin{enumerate}
    \renewcommand{\labelenumi}{(\roman{enumi})}
    \setlength{\itemsep}{2pt}
    \setlength{\topsep}{2pt}
    \item We reinterpret the RCM constraint as a rheonomic holonomic constraint and express it in a projection-based inverse-dynamics formulation, linking existing kinematic RCM descriptions to torque-level constrained dynamics in a unified framework.
    
    \item Within the constraint-consistent formulation of projected dynamics~\cite{moura2019equivalence}, we derive a torque-level control law that separates constrained and free-motion actions, enabling simultaneous RCM residual regulation, tool-tip tracking, and null-space compliance.
    
    \item We validate the resulting controller in simulation and on a RAMIS training platform against representative projection-based and constrained-dynamics baselines under spiral tracking, varying insertion depth, moving trocar conditions, and external human interaction.
\end{enumerate}

%In summary, the main contributions of this work are:  
% \begin{enumerate}[label=(\roman*)]
%     \item A reinterpretation of RCM as a rheonomic holonomic constraint embedded into a projection-based inverse-dynamics framework, reconciling task-level and kinematic formulations in a constraint-consistent~\cite{moura2019equivalence} manner.  
%     \item A torque-level controller that unifies projection-based and dynamically consistent formulations, achieving accurate tool-tip tracking, smoother and more efficient torque behavior, and reliable RCM constraint satisfaction under insertion-depth variation and trocar motion.  
%     \item Extensive validation in simulation and on a RAMIS training platform, benchmarking against two relevant category of dynamical controllers with respect RCM adaption and constraint-consistency improvement. The proposed controller is further validated under clinically relevant conditions including spiral trajectories, variable insertion depth, moving trocars, and intentional human interaction.  
% \end{enumerate}

The remainder of the paper is organized as follows: Section~\ref{sec:preliminaries} reviews the background on projection-based constrained dynamics, Section~\ref{sec:methods} presents the proposed RCM constraint kinematics and controller design, Section~\ref{sec:results} reports simulation and experimental validation, and Section~\ref{sec:conclusion} discusses limitations and concludes the work.

\section{Preliminaries} \label{sec:preliminaries}

We consider the joint-space dynamics of an \(n\in\mathbb{N}\) DoF manipulator subject to external interaction and \(k\in\mathbb{N}\) independent kinematic constraints, given by
\begin{equation}\label{eqn:inverse-dyn}
    \boldsymbol{M}\ddot{\boldsymbol{q}} + \boldsymbol{h} 
    = \boldsymbol{\tau} + \boldsymbol{J}_{c}^{T}\boldsymbol{f}_{c} + \boldsymbol{\tau}_{ext},
\end{equation}
where \(\boldsymbol{q} \in \mathbb{R}^{n}\) is the joint configuration, 
\(\boldsymbol{M} \in \mathbb{R}^{n\times n}\) is the symmetric positive-definite inertia matrix, 
\(\boldsymbol{h} \in \mathbb{R}^{n}\) collects Coriolis, centrifugal, and gravitational torques, 
\(\boldsymbol{\tau} \in \mathbb{R}^{n}\) is the control input, 
\(\boldsymbol{\tau}_{ext} \in \mathbb{R}^{n}\) is the external torque, and 
\(\boldsymbol{f}_{c} \in \mathbb{R}^{k}\) is the generalized constraint force associated with the Jacobian \(\boldsymbol{J}_{c} \in \mathbb{R}^{k\times n}\).  
The constraints satisfy \(\boldsymbol{J}_{c}\dot{\boldsymbol{q}} = \mathbf{0}\), with \(\boldsymbol{J}_{c}\) assumed to have full row rank.  

To separate constrained and free-motion dynamics, we introduce the orthogonal projector
\[
    \boldsymbol{P} = \boldsymbol{I} - \boldsymbol{J}_{c}^{\dagger}\boldsymbol{J}_{c},
\]
where \((\cdot)^{\dagger}\) denotes the Moore--Penrose pseudoinverse. 
The operator \(\boldsymbol{P}\) projects onto the null space of \(\boldsymbol{J}_{c}\) and satisfies \(\boldsymbol{P}\dot{\boldsymbol{q}} = \dot{\boldsymbol{q}}\) for all \(\dot{\boldsymbol{q}} \in \mathcal{N}(\boldsymbol{J}_{c})\) where \(\mathcal{N}(\cdot)\) defines the null-space.
It is worth noting that \(\boldsymbol{P}\) is idempotent, i.e., \(\boldsymbol{P}^{2}=\boldsymbol{P}=\boldsymbol{P}^{T}\).
Projecting \eqref{eqn:inverse-dyn} onto the free-motion subspace, it has
\begin{equation}\label{eqn:PID}
    \boldsymbol{P}\boldsymbol{M}\ddot{\boldsymbol{q}} + \boldsymbol{P}\boldsymbol{h} 
    = \boldsymbol{P}(\boldsymbol{\tau} + \boldsymbol{\tau}_{ext}).
\end{equation}
Then, we decompose the control input torque into free-motion and constrained components,
\begin{equation}\label{eqn:decompose-tau-tau}
    \boldsymbol{\tau} = \boldsymbol{\tau}_{\parallel} \oplus \boldsymbol{\tau}_{\perp},
\end{equation}
where \(\boldsymbol{\tau}_{\parallel} := \boldsymbol{P}^{\dagger}\boldsymbol{P}\boldsymbol{\tau}_{f}\) lies in the free-motion subspace,  
\(\boldsymbol{\tau}_{\perp} := (\boldsymbol{I}-\boldsymbol{P}^{\dagger}\boldsymbol{P})\boldsymbol{\tau}_{c}\) in the constrained subspace,  
and \(\boldsymbol{\tau}_{f}, \boldsymbol{\tau}_{c}\) denote free-space and constrained-space control inputs, respectively.  
The generalized inverse \(\boldsymbol{P}^{\dagger}\) is chosen dynamically consistent.  
Since \(\boldsymbol{\tau}_{\perp}\) does not contribute to motion, all free-space tasks such as trajectory tracking and null-space compliance are executed through \(\boldsymbol{\tau}_{f}\).  

For all admissible velocities \(\dot{\boldsymbol{q}} \in \mathcal{N}(\boldsymbol{J}_{c})\), it holds that \((\boldsymbol{I}-\boldsymbol{P})\dot{\boldsymbol{q}} = \mathbf{0}\).  
Differentiating with respect to time gives
\begin{equation}\label{eqn:I-P-qddot}
    (\boldsymbol{I}-\boldsymbol{P})\ddot{\boldsymbol{q}} = \dot{\boldsymbol{P}}\dot{\boldsymbol{q}},
    \qquad
    \dot{\boldsymbol{P}} = -\boldsymbol{J}_{c}^{\dagger}\dot{\boldsymbol{J}}_{c}.
\end{equation}
Combining \eqref{eqn:PID} and \eqref{eqn:I-P-qddot}, the constrained dynamics becomes
\begin{equation}\label{eqn:CPID}
    \boldsymbol{M}_{f}\ddot{\boldsymbol{q}} + \boldsymbol{P}\boldsymbol{h} - \dot{\boldsymbol{P}}\dot{\boldsymbol{q}}
    = \boldsymbol{P}(\boldsymbol{\tau} + \boldsymbol{\tau}_{ext}),
\end{equation}
where \(\boldsymbol{M}_{f} = \boldsymbol{P}\boldsymbol{M} + (\boldsymbol{I}-\boldsymbol{P})\) is nonsingular by construction.  
Pre-multiplying \eqref{eqn:CPID} with \(\boldsymbol{J}\boldsymbol{M}_{f}^{-1}\) and substituting \(\boldsymbol{J}\ddot{\boldsymbol{q}} = \ddot{\boldsymbol{x}} - \dot{\boldsymbol{J}}\dot{\boldsymbol{q}}\), the operational-space dynamics follows
\begin{equation}\label{eqn:osf-parallel}
    \boldsymbol{\Lambda}_{f}\ddot{\boldsymbol{x}} 
    + \underbrace{\boldsymbol{\Lambda}_{f}\big(\boldsymbol{J}\boldsymbol{M}_{f}^{-1}\boldsymbol{P}\boldsymbol{h} 
    - (\dot{\boldsymbol{J}} + \boldsymbol{J}\boldsymbol{M}_{f}^{-1}\dot{\boldsymbol{P}})\dot{\boldsymbol{q}}\big)}_{\boldsymbol{h}_{f}}
    = \boldsymbol{f}_{f} + \boldsymbol{f}_{ext},
\end{equation}
where \(\boldsymbol{\Lambda}_{f} = (\boldsymbol{J}\boldsymbol{M}_{f}^{-1}\boldsymbol{P}\boldsymbol{J}^{T})^{-1}\) is the task-space inertia,  
\(\boldsymbol{h}_{f}\) is the bias force, and  
\(\boldsymbol{f}_{ext} = \boldsymbol{J}^{\#T}\boldsymbol{\tau}_{ext}\) is the projected external force with \(\boldsymbol{J}^{\#T} = (\boldsymbol{J}\boldsymbol{M}_{f}^{-1}\boldsymbol{P}\boldsymbol{J}^{T})^{-1}\boldsymbol{J}\boldsymbol{M}_{f}^{-1}\boldsymbol{P}\).  
To complete a free space task, the PD+ type controller is designed,
\begin{equation}\label{eqn:constrained-OSF-F}
    \boldsymbol{f}_{f} = \boldsymbol{\Lambda}_{f}\ddot{\boldsymbol{x}}_{d} 
    + \boldsymbol{K}_{f,D}\dot{\boldsymbol{e}} + \boldsymbol{K}_{f,P}\boldsymbol{e} + \boldsymbol{h}_{f},
\end{equation}
where tracking error \(\boldsymbol{e} = \boldsymbol{x}_{d}-\boldsymbol{x}\) and positive-definite gains are \(\boldsymbol{K}_{f,P}, \boldsymbol{K}_{f,D}\).  
The corresponding joint torque input is
\begin{equation}\label{eqn:constrained-OSF-Multi}
    \boldsymbol{\tau}_{f} = \boldsymbol{J}^{T}\boldsymbol{f}_{f} + \bar{\boldsymbol{N}}\boldsymbol{\tau}_{0},
\end{equation}
where \(\bar{\boldsymbol{N}} = \boldsymbol{I} - \boldsymbol{J}^{T}\boldsymbol{J}^{\#T}\) is the dynamically consistent null-space projector, and \(\boldsymbol{\tau}_{0}\) is an auxiliary null-space input for secondary objectives lying in the free motion space.

\section{Methods} \label{sec:methods}

When virtual RCM conditions are soft or rheonomic, the separation between kinematic constraints and task-level objectives is no longer sharp. A kinematic formulation retains the structure of constraint-induced action, whereas a task-space formulation is more direct for torque-level tracking control. We therefore express the RCM in a projection-based constrained-dynamics framework~\cite{aghili2005unified} and model it as a rheonomic holonomic constraint. The resulting formulation provides a constraint-consistent torque decomposition for simultaneous RCM regulation, tool-tip tracking, and null-space compliance under trocar motion and non-ideal abdominal-wall behavior~\cite{kirilova2009visco,sadeghian2019constrained}.

\subsection{RCM Kinematics from Projection} \label{subsec:prelim-projection}

\begin{figure}[t!]
    \centering
    \vspace{5mm}
    \begin{subfigure}{0.25\textwidth}
        \includegraphics[width=\linewidth]{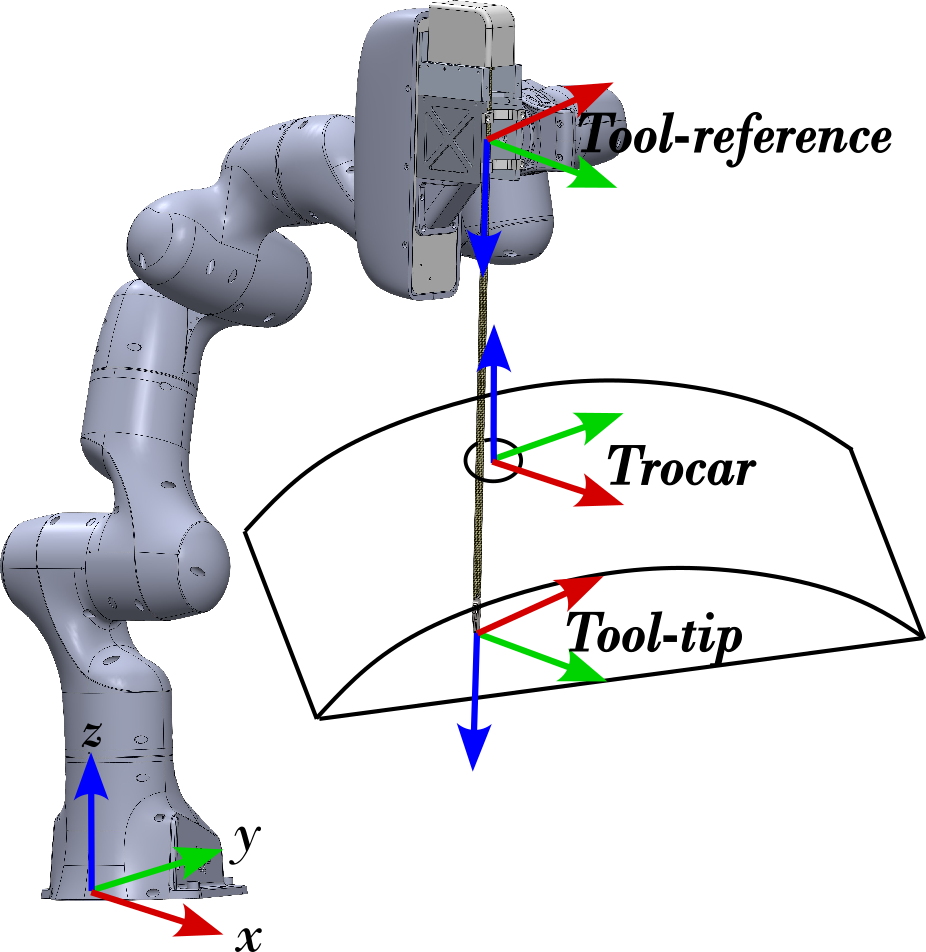}
        \caption{}
        \label{fig:robot-rcm}
    \end{subfigure}
    \hspace{-3mm}
    \begin{subfigure}{0.22\textwidth}
        \includegraphics[width=\linewidth]{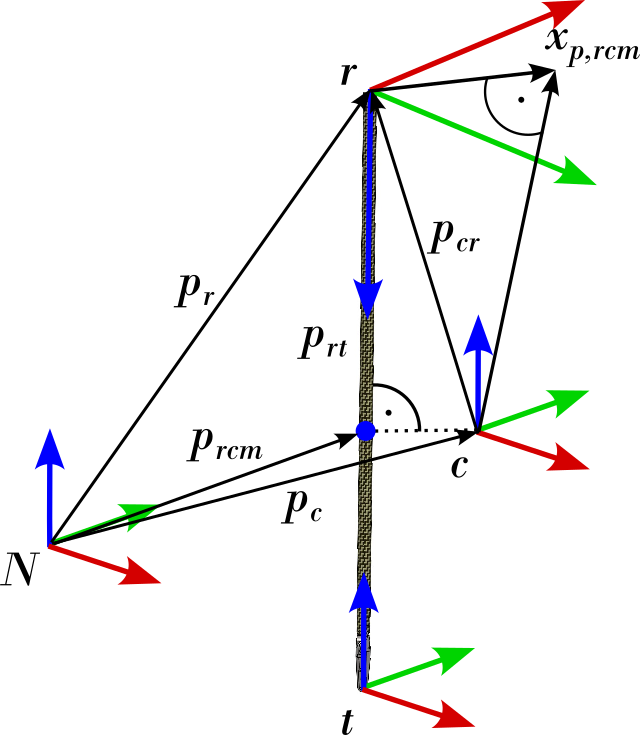}
        \caption{}
        \label{fig:robot-frame}
    \end{subfigure}
    \caption{RCM geometry and frame definitions. (a) Overview of the considered scenario: the shaft of the surgical tool (see \cite{kenanoglu2024design}) passes through the virtual RCM located at the trocar point. (b) Local frame construction used for the projected RCM residual. Frame \(r\) is attached to the tool reference point, frame \(t\) to the tool tip, and frame \(c\) to the trocar point. The residual is expressed in the \(r\)-frame to obtain the 2D/3D projected RCM coordinates used in control.}
    \label{fig:robot-rcm-frame}
    \vspace{-4mm}
\end{figure}

Let the reference pose of the surgical tool be \((\boldsymbol{p}_r,\boldsymbol{R}_r)\in\mathbb{R}^3\times\mathrm{SO}(3)\), where \(\boldsymbol{p}_r\) is the reference position and \(\boldsymbol{R}_r\) its orientation. Let the trocar point be \(\boldsymbol{p}_c(t)\), which may move with known velocity \(\dot{\boldsymbol{p}}_c(t)\). Two common formulations of RCM kinematics exist in the literature: the 3D formulation \cite{sandoval2017new} and the 2D formulation \cite{kastritsi2021control}. To prepare for our later projection-based controller, we reinterpret both approaches by expressing them in the local tool-reference frame (the \(r\)-frame), as detailed in Fig.~\ref{fig:robot-rcm-frame} with RCM kinematics. This ensures consistency with the projection formulation introduced in Section~\ref{sec:preliminaries}, while maintaining equivalence to the original derivations.

The projected residual is defined as the orthogonal projection of the residual vector onto the \(r\)-frame. This approach adapts the idea presented in \cite{sandoval2017new} by formulating the projection in the \(r\)-frame rather than the base frame.
\begin{equation}\label{eqn:3D-rcm-error}
    \boldsymbol{x}_{p,rcm,3D} = \boldsymbol{R}_r^{T} \boldsymbol{p}_{cr} \in \mathbb{R}^3
\end{equation}
with \(\boldsymbol{p}_{cr} = \boldsymbol{p}_r - \boldsymbol{p}_c\). The associated Jacobian with respect to \(\boldsymbol{p}_c\) is
\begin{equation}
\begin{bmatrix}
\boldsymbol{J}_{p,c}\\[2pt]
\boldsymbol{J}_{\omega,c}
\end{bmatrix}
=
\begin{bmatrix}
\boldsymbol{I}_{3\times3} & -\boldsymbol{p}_{cr}^{\wedge}\\
\boldsymbol{0}_{3\times3} & \boldsymbol{I}_{3\times3}
\end{bmatrix}
\begin{bmatrix}
\boldsymbol{J}_{p,r}\\[2pt]
\boldsymbol{J}_{\omega,r}
\end{bmatrix},
\end{equation}
where \(\boldsymbol{J}_{p,r}\) and \(\boldsymbol{J}_{\omega,r}\) are the translational and angular parts of the Jacobian at the tool reference, while \(\boldsymbol{J}_{p,c}\) and \(\boldsymbol{J}_{\omega,c}\) correspond to the trocar point. The RCM Jacobian then follows the orthogonal projection correspondingly, i.e.,
\begin{equation} \label{eqn:3D-rcm-jacobian}
    \boldsymbol{J}_{rcm,3D} = \boldsymbol{R}_r^{T} \boldsymbol{J}_{c,p}.
\end{equation}
Consequently, the projected velocity residual becomes
\begin{equation}\label{eqn:3d-rcm-error-derivative}
    \dot{\boldsymbol{x}}_{p,rcm,3D} = \boldsymbol{J}_{rcm,3D}\dot{\boldsymbol{q}} - \boldsymbol{R}_r^{T}\dot{\boldsymbol{p}}_c.
\end{equation}

Notably, for the 2D case following \cite{kastritsi2021control}, we define a planar basis \(\boldsymbol{B}_r \in \mathbb{R}^{3 \times 2}\) by taking the first two columns of \(\boldsymbol{R}_r\). Replacing \(\boldsymbol{R}_r\) with \(\boldsymbol{B}_r\) in \eqref{eqn:3D-rcm-error}--\eqref{eqn:3d-rcm-error-derivative}, we obtain the corresponding Jacobian \(\boldsymbol{J}_{rcm,2D}\), and 2D residual \({\boldsymbol{x}}_{p,rcm,2D}\) and velocity \(\dot{\boldsymbol{x}}_{p,rcm,2D}\).

\subsection{Constraint-consistent control} \label{sec:rcm-control}

Non-ideal kinematic or task-level constraints can be described by general rheonomic holonomic conditions, i.e.,
\begin{align}
\label{eq_constraint}
    \boldsymbol{\Phi}_{c}(\boldsymbol{q},t) = \boldsymbol{x}_{c}(t) \in \mathbb{R}^{k},
\end{align}
where \(\boldsymbol{x}_{c}(t)\) denotes the time varying residual constraint coordinate. \((\cdot)_{c}\) represents not only the placeholder of different RCM constraint formulations but also for any generic constraints.  
To obtain the velocity and acceleration relationships, we differentiate the constraint equation \eqref{eq_constraint} with respect to time as follows
\begin{equation}
    \frac{\partial \boldsymbol{\Phi}_{c}}{\partial \boldsymbol{q}} \dot{\boldsymbol{q}} 
    + \frac{\partial \boldsymbol{\Phi}_{c}}{\partial t} = \dot{\boldsymbol{x}}_{c}, 
    \quad
    \frac{\partial^2 \boldsymbol{\Phi}_{c}}{\partial \boldsymbol{q}^2}\dot{\boldsymbol{q}} 
    + \frac{\partial \boldsymbol{\Phi}_{c}}{\partial \boldsymbol{q}}\ddot{\boldsymbol{q}} 
    + \frac{\partial^2 \boldsymbol{\Phi}_{c}}{\partial t^2} = \ddot{\boldsymbol{x}}_{c}.
\end{equation}
For brevity, we define the Jacobian of $\boldsymbol{x}_c$ as \(\boldsymbol{J}_{c} = \partial \boldsymbol{\Phi}_{c} / \partial \boldsymbol{q}\) and the lumped nonlinear bias term at acceleration level \(\boldsymbol{b}_{c} := \tfrac{\partial^2 \boldsymbol{\Phi}_{c}}{\partial \boldsymbol{q}^2}\dot{\boldsymbol{q}} + \tfrac{\partial^2 \boldsymbol{\Phi}_{c}}{\partial t^2}\) \footnote{For brevity, the computation is not analytically elaborated with respect to different RCM or any constraint implementations.}.  
We then obtain the acceleration-level equivalence subject to constraint 
\begin{equation} \label{eqn:acc-constraint}
    \boldsymbol{J}_{c}\ddot{\boldsymbol{q}} = \ddot{\boldsymbol{x}}_{c} - \boldsymbol{b}_{c}.
\end{equation}
Pre-multiplying by \(\boldsymbol{J}_{c}^{\dagger}\) gives
\begin{equation}
    (\boldsymbol{I}-\boldsymbol{P})\ddot{\boldsymbol{q}} 
    = \boldsymbol{J}_{c}^{\dagger}(\ddot{\boldsymbol{x}}_{c} - \boldsymbol{b}_{c}).
\end{equation} 
Thus, the joint acceleration decomposes as
\[
\ddot{\boldsymbol{q}} = \ddot{\boldsymbol{q}}_{\parallel} \oplus \ddot{\boldsymbol{q}}_{\perp},
\]
with \(\ddot{\boldsymbol{q}}_{\parallel} = \boldsymbol{P}\ddot{\boldsymbol{q}}\) and \(\ddot{\boldsymbol{q}}_{\perp} = (\boldsymbol{I}-\boldsymbol{P})\ddot{\boldsymbol{q}}\).  
Substituting the dynamics \eqref{eqn:inverse-dyn}, we obtain
\[
\ddot{\boldsymbol{q}} = \boldsymbol{J}_{rcm}^{\dagger}(\ddot{\boldsymbol{x}}_{c} - \boldsymbol{b}_{c}) 
+ \boldsymbol{P}\boldsymbol{M}^{-1}(\boldsymbol{\tau} + \boldsymbol{\tau}_{ext} - \boldsymbol{h}),
\]
which combines constrained and unconstrained dynamics in the sense of Gauss's principle \cite{udwadia2002foundations}.  
Accordingly, \eqref{eqn:CPID} updates to
\begin{equation} \label{eqn:joint-space-dyn-rcm}
    \boldsymbol{M}_{f}\ddot{\boldsymbol{q}} 
    = \boldsymbol{P}(\boldsymbol{\tau}_{f} + \boldsymbol{\tau}_{ext} - \boldsymbol{h})
    - \boldsymbol{J}_{rcm}^{\dagger}(\ddot{\boldsymbol{x}}_{c} - \boldsymbol{b}_{c}).
\end{equation}

The corresponding operational formulation \eqref{eqn:osf-parallel} is obtained by substituting
\begin{equation} \label{eqn:h-parallel-rcm}
    \boldsymbol{h}_{f} = \boldsymbol{\Lambda}_{f}\big(
        \boldsymbol{J}\boldsymbol{M}_{f}^{-1}\boldsymbol{P}\boldsymbol{h}
        - \dot{\boldsymbol{J}}\dot{\boldsymbol{q}}
        + \boldsymbol{J}\boldsymbol{M}_{f}^{-1}\boldsymbol{J}_{rcm}^{\dagger}(\ddot{\boldsymbol{x}}_{c} - \boldsymbol{b}_{c})
    \big).
\end{equation}
Hence, the operational-space control law \eqref{eqn:constrained-OSF-F} applies with the updated \(\boldsymbol{h}_{f}\).  

To stabilize the residual dynamics, the constraint acceleration is designed as
\begin{equation} \label{eqn:xc_ddot_design}
    \ddot{\boldsymbol{x}}_{c} = \ddot{\boldsymbol{x}}_{c,d} 
    - \boldsymbol{\Lambda}_{c}^{-1}(
        \boldsymbol{K}_{c,D}\dot{\tilde{\boldsymbol{x}}}_{c} 
        + \boldsymbol{K}_{c,P}\tilde{\boldsymbol{x}}_{c}
    ) + \boldsymbol{b}_{c},
\end{equation}
where \(\tilde{\boldsymbol{x}}_{c} = \boldsymbol{x}_{c}-\boldsymbol{x}_{c,d}\) denotes the constraint error,  
\(\boldsymbol{K}_{c,P}, \boldsymbol{K}_{c,D} \succ 0\) are gain matrices, and  
\(\boldsymbol{\Lambda}_{c} = (\boldsymbol{J}_{c}\boldsymbol{M}^{-1}\boldsymbol{J}_{c}^{T})^{-1}\) is the constraint-space inertia. For most cases (e.g. RCM constraints), \(\boldsymbol{x}_{c,d}=\boldsymbol{0}\) will be defined for exact satisfaction.

The associated constrained torque in \eqref{eqn:decompose-tau-tau} is then designed as
\begin{equation}\label{eqn:tau-perp}
    \boldsymbol{\tau}_{c} = \boldsymbol{J}_{c}^{T}(
        \boldsymbol{\Lambda}_{c}\ddot{\boldsymbol{x}}_{c}
        + \boldsymbol{J}_{c}\boldsymbol{M}^{-1}\boldsymbol{h}
    ).
\end{equation}
In analogy to the free-space formulation, this ensures that \(\boldsymbol{x}_{p,rcm}\) and its derivatives converge to zero.  
Any disturbance induced by the additional constraint dynamics is compensated by updating \(\ddot{\boldsymbol{x}}_{c}\) via \eqref{eqn:xc_ddot_design}.

\textbf{Remark 1.} \textit{To enhance robustness, we extend the control law with a disturbance estimate \(\hat{\boldsymbol{\tau}}_{ext}\), such that}
\begin{equation} \label{eqn:tau+tau-ext}
    \hat{\boldsymbol{\tau}} = \boldsymbol{\tau}_{\parallel} \oplus \boldsymbol{\tau}_{\perp} + \hat{\boldsymbol{\tau}}_{ext}.
\end{equation}
\textit{The estimate \(\hat{\boldsymbol{\tau}}_{ext}\) can be decomposed into free and constrained components using \(\boldsymbol{P}\) and is obtained via a momentum-based observer with sign inversion, so that \(\hat{\boldsymbol{\tau}}_{ext} + \boldsymbol{\tau}_{ext} \to 0\) \cite{de2006collision}.  
Finally, null-space compliance is introduced through}
\begin{equation} \label{eqn:tau-nullimp}
    \boldsymbol{\tau}_{0} = -\boldsymbol{K}_{n,D}\dot{\tilde{\boldsymbol{q}}} - \boldsymbol{K}_{n,P}\tilde{\boldsymbol{q}},
\end{equation}
\textit{where \(\tilde{\boldsymbol{q}} = \boldsymbol{q}-\boldsymbol{q}_{init}\) is defined relative to the initial configuration, and \(\boldsymbol{K}_{n,P}, \boldsymbol{K}_{n,D} \succ 0\) are null-space impedance gains.  
The final control input \(\hat{\boldsymbol{\tau}}\) thus replaces \(\boldsymbol{\tau}\) in \eqref{eqn:inverse-dyn} under disturbances.}

\subsection{Baseline of projection controllers for comparison} \label{sec:ap-a}

\subsubsection{Projection Jacobian approach} \label{secsec:Z-projection-approach}

For comparison, we first outline an alternative operational-space formulation underlying the admittance controller in~\cite{kastritsi2021control}. The dynamics in operational space are expressed as
\begin{align}
    \boldsymbol{\Lambda}_E\ddot{\boldsymbol{x}} + \boldsymbol{H}_E 
    = \begin{bmatrix}
        \boldsymbol{f}_{c} \\
        \boldsymbol{f}_{n}
    \end{bmatrix},
\end{align}
where
\begin{align}
    \boldsymbol{\Lambda}_E &= \boldsymbol{J}_E^{-T}\boldsymbol{M}\boldsymbol{J}_E^{-1} 
    = \begin{bmatrix}
        \boldsymbol{\Lambda}_{c} & 0 \\
        0 & \boldsymbol{\Lambda}_{n} 
    \end{bmatrix}, \\
    \boldsymbol{\Lambda}_{c} &= (\boldsymbol{J}_{c}\boldsymbol{M}^{-1}\boldsymbol{J}_{c}^{T})^{-1}, \\
    \boldsymbol{\Lambda}_{n} &= \boldsymbol{Z}^{T}\boldsymbol{M}\boldsymbol{Z}, \\
    \boldsymbol{H}_E &= \boldsymbol{\Lambda}_E\left(
    \begin{bmatrix}
        \boldsymbol{J}_{c} \\
        \boldsymbol{Z}^{\#}
    \end{bmatrix} \boldsymbol{M}^{-1}\boldsymbol{h} - 
    \begin{bmatrix}
        \dot{\boldsymbol{J}}_{c} \\
        \tfrac{d}{dt}\boldsymbol{Z}^{\#}
    \end{bmatrix}\dot{\boldsymbol{q}} \right).
\end{align}

The extended Jacobian \(\boldsymbol{J}_E\) is defined by
\begin{equation}
    \begin{bmatrix}
        \dot{\boldsymbol{x}}_{c} \\
        \dot{\boldsymbol{x}}_{n}
    \end{bmatrix} = 
    \underbrace{\begin{bmatrix}
        \boldsymbol{J}_{c} \\
        \boldsymbol{Z}^{\#}
    \end{bmatrix}}_{\boldsymbol{J}_E}\dot{\boldsymbol{q}},
\end{equation}
where \(\dot{\boldsymbol{x}}_{n}\) denotes the null-space velocity orthogonal to the constraint space, and \(\boldsymbol{Z}^{\#} = (\boldsymbol{Z}^{T}\boldsymbol{M}\boldsymbol{Z})^{-1}\boldsymbol{Z}^{T}\boldsymbol{M}\).  
The corresponding torque input is
\begin{equation}
    \boldsymbol{\tau} = \boldsymbol{J}_{E}^{T} 
    \left(\begin{bmatrix}
        \boldsymbol{f}_{c} \\
        \boldsymbol{Z}^{T}\boldsymbol{J}^{T}\boldsymbol{f}_{f}
    \end{bmatrix} + \boldsymbol{H}_E\right) + \hat{\boldsymbol{\tau}}_{ext},
\end{equation}
where the constraint \(\boldsymbol{JZ} = \mathbf{0}\) holds and \(\boldsymbol{Z}\) is chosen following \cite{ott2008cartesian}.  
For simplicity and comparability, we restrict to the static RCM case with \(\dot{\boldsymbol{p}}_{c} = \mathbf{0}\). The constrained force \(\boldsymbol{f}_c\) is implemented through a PD law relative to \(-\boldsymbol{x}_{c}\),
\begin{equation}
    \boldsymbol{f}_{c} = \boldsymbol{\Lambda}_{c}\ddot{\boldsymbol{x}}_{c,d} 
    -\boldsymbol{K}_{c,D}\dot{\tilde{\boldsymbol{x}}}_{c} - \boldsymbol{K}_{c,P}\tilde{\boldsymbol{x}}_{c},
\end{equation}
The free-space force \(\boldsymbol{f}_f\) is implemented analogously to \eqref{eqn:constrained-OSF-F}.

\subsubsection{Dynamical Udwadia--Kalaba controller} \label{secsec:U-K-controller}

We also reformulate an existing torque controller \cite{minelli2022torque}, originally derived from a dynamically consistent framework, but lacking explicit constraint consistency in the residual dynamics. Starting from the constraint relation \eqref{eqn:acc-constraint}, the inverse dynamics are decomposed according to the Udwadia--Kalaba (U--K) principle \cite{udwadia2002foundations}:
\begin{equation}
    \boldsymbol{M}\ddot{\boldsymbol{q}} 
    = \boldsymbol{Q} + \boldsymbol{Q}_{ic} + \boldsymbol{Q}_{nic} + \boldsymbol{h} + \boldsymbol{\hat{\tau}}_{ext},
\end{equation}
where \(\boldsymbol{Q}\), \(\boldsymbol{Q}_{ic}\), and \(\boldsymbol{Q}_{nic}\) denote the contributions from free motion, ideal constraints, and non-ideal constraints, respectively:
\begin{align}
    \boldsymbol{Q} &= \boldsymbol{\tau}^{\#} - \boldsymbol{h}, \\
    \boldsymbol{Q}_{ic} &= \boldsymbol{M}^{1/2} \boldsymbol{\Pi}^{\dagger}(\boldsymbol{b}_{ic} - \boldsymbol{J}_{c}\boldsymbol{M}^{-1}\boldsymbol{Q}), \\
    \boldsymbol{Q}_{nic} &= \boldsymbol{M}^{1/2}\!\left[\boldsymbol{I} - \boldsymbol{\Pi}^{\dagger}\boldsymbol{J}_{c}\boldsymbol{M}^{-1/2}\right]\!\boldsymbol{M}^{-1/2}\boldsymbol{\tau}_{nic},
\end{align}
with \(\boldsymbol{\tau}^{\#}\) the free-space torque (designed as Cartesian PD+ with optional null-space term), \(\boldsymbol{\Pi} = \boldsymbol{J}_{c}\boldsymbol{M}^{-1/2}\), and
\begin{equation}
    \boldsymbol{\tau}_{nic} = \boldsymbol{J}_{c}^{T}\left(
        \boldsymbol{\Lambda}_{c}\ddot{\boldsymbol{x}}_{c,d} 
        - \boldsymbol{K}_{c,D}\dot{\tilde{\boldsymbol{x}}}_{c} 
        - \boldsymbol{K}_{c,P}\tilde{\boldsymbol{x}}_{c}
    \right),
\end{equation}
as in \eqref{eqn:tau-perp}.  
Similarly to the proposed controller, the term 
\(
\boldsymbol{b}_{ic} = \boldsymbol{\Lambda}_{c}\ddot{\boldsymbol{x}}_{c,d} 
- \boldsymbol{K}_{c,D}\dot{\tilde{\boldsymbol{x}}}_{c} 
- \boldsymbol{K}_{c,P}\tilde{\boldsymbol{x}}_{c}
\)
accounts for ideal rheonomic holonomic constraints and compensates for the part from non-ideal constraints
minimization efforts \(\boldsymbol{\tau}_{nic}\).  
Unlike \cite{minelli2022torque}, where \(\boldsymbol{\tau}_{nic}\) is simply added to \(\boldsymbol{Q}\) and \(\boldsymbol{Q}_{ic}\), here the non-ideal constraint is consistently projected into torque space, leading to a self-contained constraint-consistent formulation.

\begin{figure*}[h!]
    \centering % Center the entire figure content

    % The first sub-figure
    \begin{subfigure}[b]{0.9\textwidth}
        \centering
        \includegraphics[width=\textwidth]{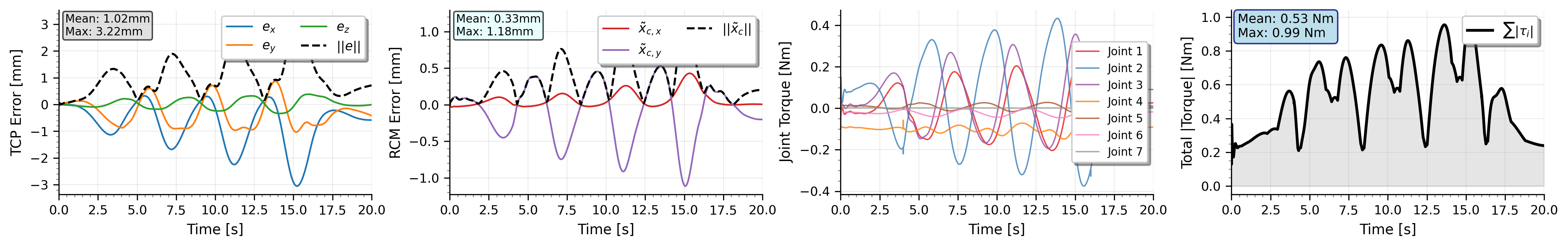}
        \caption{Proposed controller: \(\boldsymbol{P}\)-approach}
        \label{fig:sim-proposed}
    \end{subfigure}
    \hfill 
    \begin{subfigure}[b]{0.9\textwidth}
        \centering
        \includegraphics[width=\textwidth]{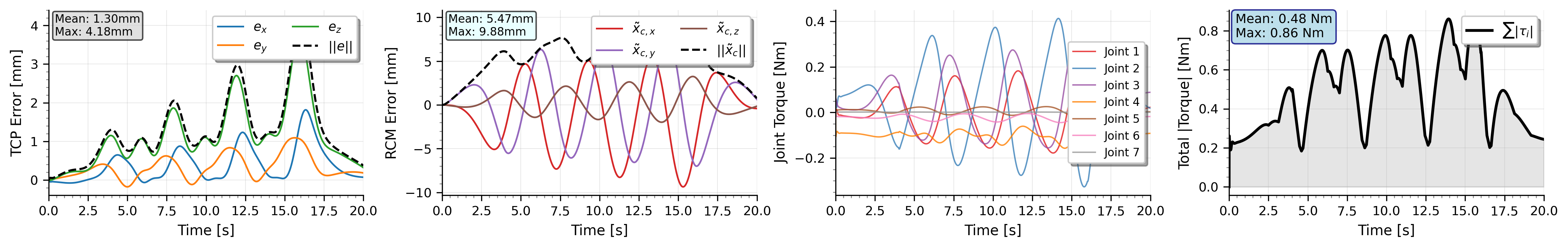}
        \caption{U-K Approach}
        \label{fig:sim-minelli}
    \end{subfigure}
    \hfill 
    \begin{subfigure}[b]{0.9\textwidth}
        \centering
        \includegraphics[width=\textwidth]{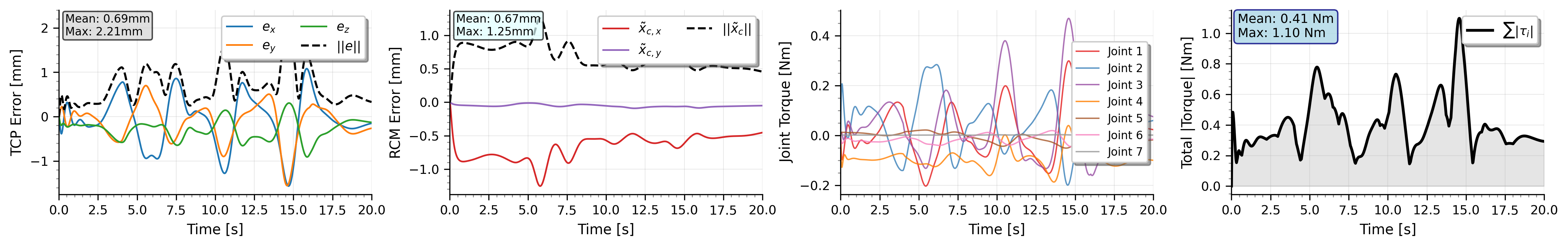}
        \caption{\(\boldsymbol{Z}\)-approach}
        \label{fig:sim-z-projection}
    \end{subfigure}
    \caption{Simulation comparison of the three controllers. For each controller, the panels report tool-tip tracking error, RCM residual, and joint-torque evolution along the same spiral reference trajectory. (a) Proposed \(\boldsymbol{P}\)-approach. (b) U--K approach. (c) \(\boldsymbol{Z}\)-approach.}
    \label{fig:mujoco-sim-compared}
\end{figure*}

\section{Results} \label{sec:results}
In this section, the proposed controller is first validated through simulation and experiments in comparison with baseline controllers. Subsequently, we examine its performance in dynamic and human-interactive tasks to demonstrate reliability in practice.

\subsection{Task implementation} \label{sec:task-implementation}
For benchmarking, the controllers are implemented following their original formulations. The primary task is tool-tip position tracking, while the constraint task is RCM regulation. The U--K approach of Minelli et al.~\cite{minelli2022torque} is evaluated under the 3D RCM formulation, whereas the proposed controller and the \(\boldsymbol{Z}\)-approach adopt the 2D projected formulation. Accordingly, \(\boldsymbol{x}_{c,d}\) and its derivatives are set to zero, and all terms \((\cdot)_c\) are instantiated as \((\cdot)_{p,rcm,3D}\) or \((\cdot)_{p,rcm,2D}\), respectively.

A dynamically challenging spiral trajectory \(\boldsymbol{p}_{t,d}(t)\) is generated along the robot \(z\)-axis. The trajectory has a radius of \(r=0.02\,\mathrm{m}\), a pitch of \(0.015\,\mathrm{m}\), and a trapezoidal velocity profile over a duration of \(T=20\,\mathrm{s}\). The free-motion task is defined by \(\boldsymbol{x}_d=\boldsymbol{p}_{t,d}\) and \(\boldsymbol{x}=\boldsymbol{p}_t\), together with their corresponding derivatives. In the experiments, the reported tip-tracking error and projected RCM residual are reconstructed from the robot state and calibrated tool geometry, and therefore reflect both control performance and residual modeling mismatch.

The trocar position \(\boldsymbol{p}_{c}\) is initialized from the starting configuration \((\boldsymbol{p}_{r}(0),\boldsymbol{p}_{t}(0))\) using a scaling factor \(\alpha\):
\begin{equation}
    \boldsymbol{p}_{c} = \boldsymbol{p}_{r}(0) + \alpha \left(\boldsymbol{p}_{t}(0) - \boldsymbol{p}_r(0) \right).
\end{equation}
The physical insertion depth is \(\alpha \cdot L_{\text{tool}}\), where the tool length is \(L_{\text{tool}} = 0.59\,\mathrm{m}\). A larger \(\alpha\) corresponds to a shallower insertion.

In practice, trocar motion can be induced by breathing or other tissue motion~\cite{sadeghian2019constrained}. To emulate this effect, a sinusoidal motion of \(\boldsymbol{p}_c\) along the \(z\)-axis with frequency \(0.2\,\mathrm{Hz}\) and amplitude \(\pm 4\,\mathrm{cm}\) is predefined in the experiments.

\subsection{Performance validation through simulation}
To validate the proposed constraint-consistent controller (\(\boldsymbol{P}\)-approach), we compare it with two representative baselines: the projection Jacobian controller (\(\boldsymbol{Z}\)-approach, Sec.~\ref{secsec:Z-projection-approach}) and the dynamical U--K controller (Sec.~\ref{secsec:U-K-controller}).

A MuJoCo simulation environment is built using the same robot kinematics, tool geometry, trocar initialization, and spiral reference described in Sec.~\ref{sec:task-implementation}, with a timestep of \(1\,\mathrm{ms}\). To emulate non-ideal contact around the trocar region, compliant MuJoCo soft objects are introduced with tuned impedance parameters allowing limited penetration. For simplicity, rheonomic trocar motion is omitted in simulation and \(\boldsymbol{p}_c\) is kept fixed. Gains are selected such that the proposed \(\boldsymbol{P}\)-approach and the \(\boldsymbol{Z}\)-approach achieve comparable tip-tracking and RCM-regulation levels, while the same gains are applied to the U--K controller for consistency; derivative gains are set element-wise as \(k_{D,i}=2\sqrt{k_{P,i}}\).

Compared with the \(\boldsymbol{Z}\)-approach, the proposed \(\boldsymbol{P}\)-approach requires slightly higher mean absolute torque (0.53 vs.\ 0.41\,Nm) but yields a lower peak torque (0.99 vs.\ 1.10\,Nm) and smoother joint-torque profiles over the periodic spiral motion. By contrast, the \(\boldsymbol{Z}\)-approach shows more irregular torque transients despite comparable tracking and RCM regulation. Relative to the U--K controller, the proposed method also yields lower tip-tracking and RCM residuals in the tested gain range. When implemented in our setup with the selected gains, the original U--K baseline became numerically unstable if only the additional \(\boldsymbol{\tau}_{nic}\) term from~\cite{minelli2022torque} was added.

\subsection{Performance validation through experiment}

\subsubsection{Comparative evaluation of the proposed controller}
Experiments are conducted on a Franka Research 3 (FR3) robot arm with a real-time loop running at 1\,kHz under Ubuntu~22.04 with a real-time kernel.  

\begin{figure}[t!]
    \centering
    \includegraphics[width=1\linewidth]{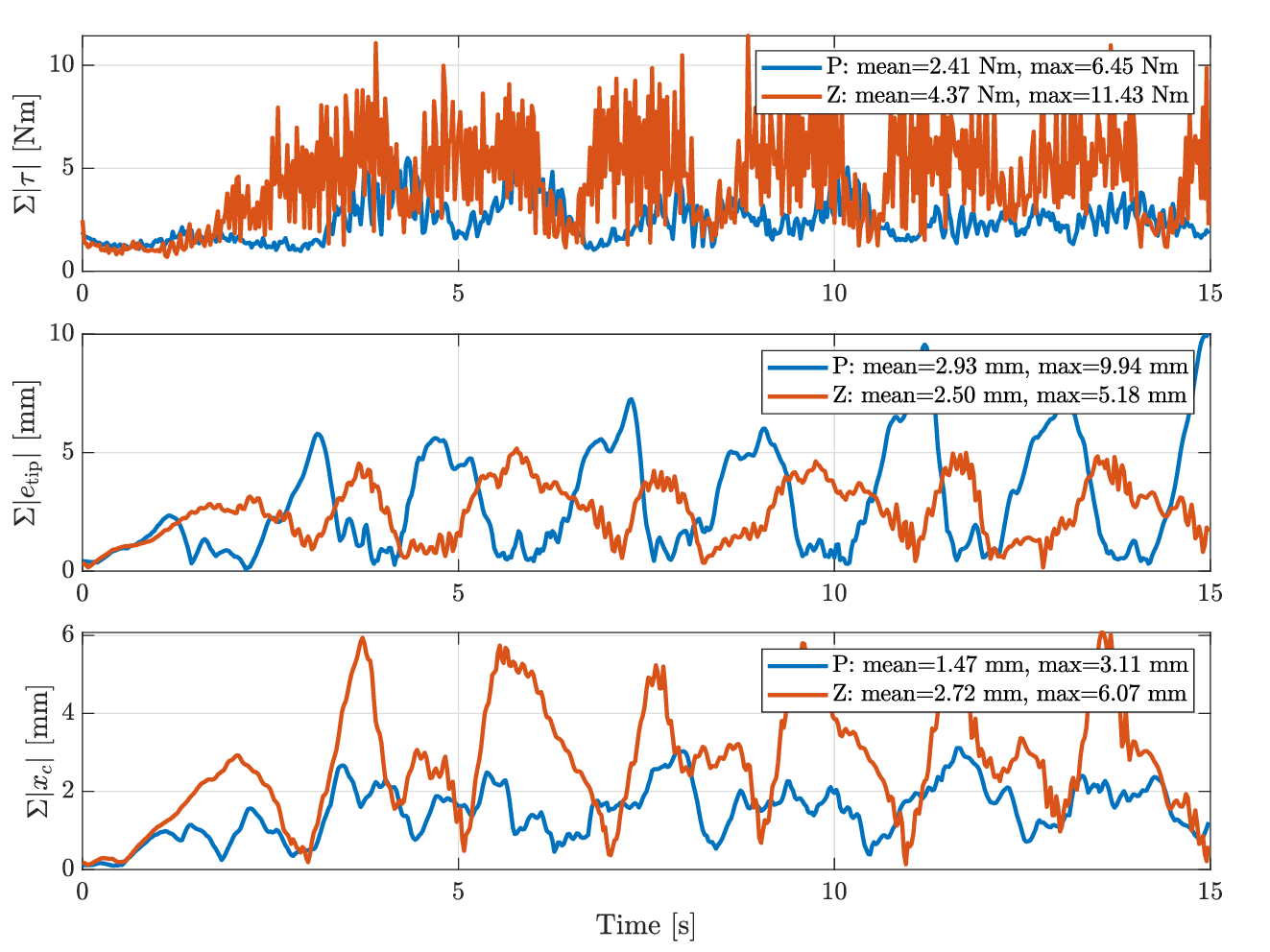}
    \caption{Experimental comparison between the proposed controller and the \(\boldsymbol{Z}\)-approach under comparable tracking and RCM residual levels. The figure reports joint-torque evolution together with the corresponding tip-tracking and RCM residual signals.}
    \label{fig:P-Z-compare}
    \vspace{-5mm}
\end{figure}

To further validate the simulation results, we compare the proposed controller (\(\boldsymbol{P}\)-approach) with the \(\boldsymbol{Z}\)-approach. Both controllers are evaluated on the same spiral reference, with the same trocar initialization and insertion depth, and under the same tool configuration. The U--K approach could not be stably implemented with the same or comparable gains as the \(\boldsymbol{P}\)-approach, likely due to numerical issues arising from heavy matrix manipulations and sensitivity to modeling errors. The gains of the \(\boldsymbol{Z}\)-approach are tuned in the same manner as in simulation\footnote{For transparency, the proportional gains applied to the tool tip and RCM tasks are 1000 and 1500~N/m, respectively, in diagonal form.}, and the tracking error trajectories and statistics are shown in Fig.~\ref{fig:P-Z-compare}.  

The proposed controller achieves lower RCM residuals while showing slightly weaker tip tracking than the \(\boldsymbol{Z}\)-approach. Nevertheless, the total torque consumption of the \(\boldsymbol{P}\)-approach, computed as the sum of absolute joint torques, is about half of that of the \(\boldsymbol{Z}\)-approach. Moreover, the peak torques of the \(\boldsymbol{Z}\)-approach are 77\% higher than those of the proposed controller. This reduction is consistent with the orthogonal decomposition of constrained and free-motion torques in the constraint-consistent formulation. In particular, the decomposition separates the control action into projected free-motion and constraint-related components, in line with minimum-effort arguments commonly used in projected inverse-dynamics formulations~\cite{mistry2012operational}. Additionally, the \(\boldsymbol{Z}\)-approach produces more aggressive torque trajectories, consistent with the non-smooth behavior observed in simulation due to the lack of a dynamical constraint-consistent formulation, which likely amplifies overall torque consumption.
Projection-Jacobian controllers are known to admit passivity-related issues in hierarchical settings~\cite{dietrich2015passivation}. The proposed controller differs structurally by embedding the RCM regulation into a constraint-consistent projected dynamics formulation rather than a classical projected subordinate compliance law; however, a formal passivity proof for the present rheonomic formulation is left for future work.

\begin{figure*}[t!]
    \centering
    \begin{subfigure}[b]{0.9\textwidth}
        \centering
        \includegraphics[width=\textwidth, trim=50 0 50 0,clip]{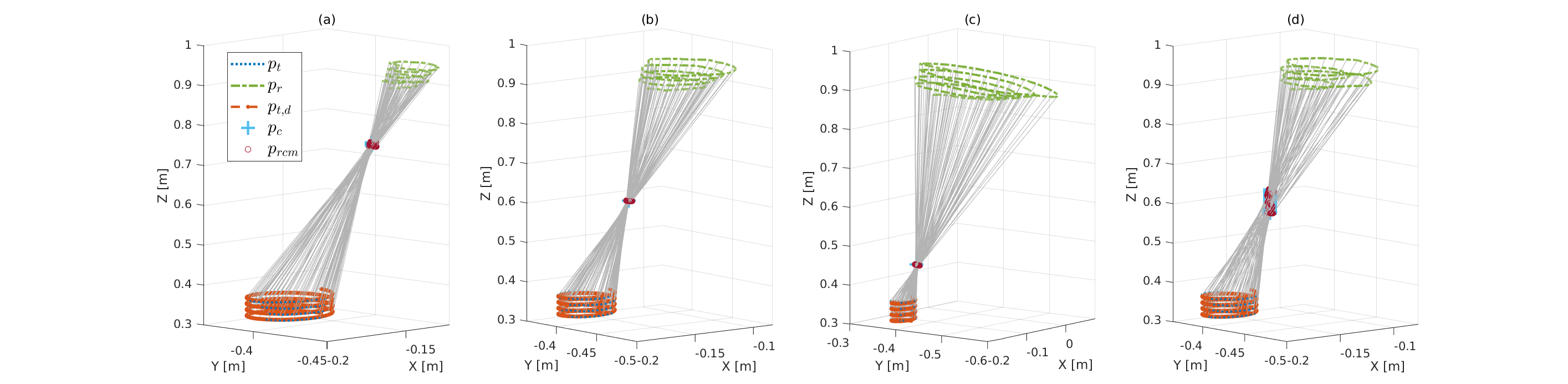}
    \end{subfigure}
    \hfill 
    \begin{subfigure}[b]{0.9\textwidth}
        \centering
        \includegraphics[width=\textwidth, trim=50 0 50 0,clip]{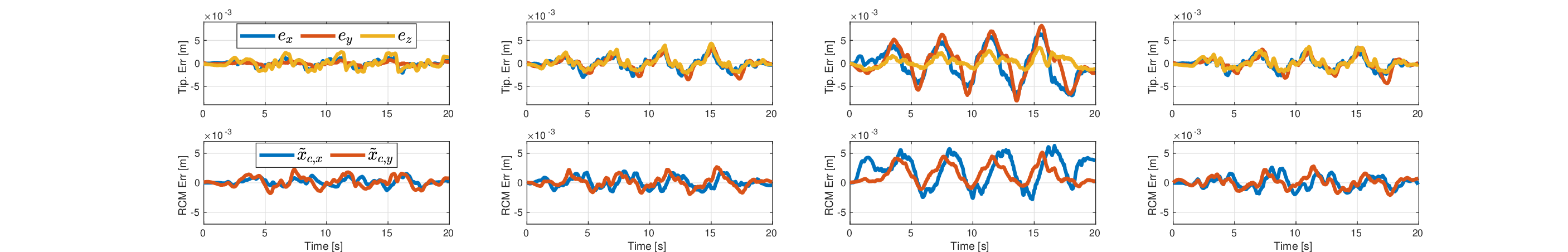}
    \end{subfigure}
    \caption{Experimental results of the proposed controller under varying insertion depth and rheonomic trocar motion. Columns (a)--(c) correspond to insertion factors \(\alpha=0.75\), \(0.50\), and \(0.25\), respectively; column (d) corresponds to the moving-trocar case. From top to bottom, the rows show 3D task-space visualization, tool-tip tracking error, and projected RCM residual.}
    \label{fig:insertion-exp}
    % \vspace{-7mm}
\end{figure*}

\subsubsection{Dynamical task feasibility experiment}
The insertion depth varies with the surgical objective and can be adjusted either by semi-autonomous guidance of the torque-controlled manipulator or by teleoperation. To investigate this, experiments are conducted at three insertion depths, corresponding to scaling factors \(\alpha = 75\%, 50\%, 25\%\). The results are visualized in Fig.~\ref{fig:insertion-exp}(a--c).  

For visualization only, the instantaneous RCM point in the base frame is reconstructed following~\cite{sandoval2017new} as
\[
    \boldsymbol{p}_{rcm} = \boldsymbol{p}_r + \frac{1}{L_{tool}^2}\boldsymbol{p}_{rt}^{T} \boldsymbol{p}_{rc} \boldsymbol{p}_{rt}
\]
where \(\boldsymbol{p}_{rt} = \boldsymbol{p}_t - \boldsymbol{p}_r\) with the tool-tip position \(\boldsymbol{p}_t\) expressed in the base frame (Fig.~\ref{fig:robot-frame}). The quantitative evaluation in Table~\ref{tab:error_cases} is based on the projected RCM residual introduced in Sec.~\ref{subsec:prelim-projection}.

The results displayed in Fig. 5(a--c) show good spatial correspondence between the desired \(\boldsymbol{p}_c\) (blue cross) and the reconstructed \(\boldsymbol{p}_{rcm}\) values (magenta circles), together with small projected residuals. However, shallower insertions (smaller \(\alpha\)) make both tip tracking and RCM constraint satisfaction more difficult. The tool-reference motion \(\boldsymbol{p}_{r}\) (green trace) increases by roughly a factor of three from case~(a) to case~(c). This lever-arm effect means that shallower insertion amplifies lateral displacements for the same angular deviation. Consequently, the system undergoes faster motions, which magnify errors caused by model mismatches, in particular due to discrepancies in the customized surgical tool drive unit. Consistently, the tip error also scales with the lever arm, being about three times larger in case~(c) than in case~(a). The quantitative results are summarized in Table~\ref{tab:error_cases}.  

Finally, rheonomic effects are evaluated by activating trocar motion as described in Sec.~\ref{sec:task-implementation}. The numerical error in Fig.~\ref{fig:insertion-exp}(d) and the mean absolute error in Table~\ref{tab:error_cases} remain small and close to those of case~(c) under the same insertion depth. This indicates that the rheonomic constraints are handled consistently by the proposed controller in the tested setting.

\begin{table}[t!]
    \centering\scriptsize
    \vspace{2mm}
    \caption{Mean absolute Cartesian tip-tracking error and projected RCM residual for the four experimental cases. Case (a): \(\alpha=0.75\); case (b): \(\alpha=0.50\); case (c): \(\alpha=0.25\); case (d): moving trocar. Units are \(10^{-3}\,\mathrm{m}\).}
    \label{tab:error_cases}
    \renewcommand{\arraystretch}{1.1} % Adjust row spacing for readability
    \begin{tabular}{c c c c c c}
        \toprule
        \textbf{Case} & \multicolumn{3}{c}{\textbf{Tip Tracking Error \(\boldsymbol{e}_{(x, y, z)}\) }} & \multicolumn{2}{c}{\textbf{Projected RCM Residual \(\tilde{\boldsymbol{x}}_{c,{(x, y)}}\)}}\\
        \midrule
        a & 0.5066 & 0.2504 & 0.8444 & 0.4152 & 0.6824 \\
        b & 0.9992 & 0.9972 & 0.9155 & 0.7336 & 0.7079\\
        c & 3.0231 & 3.0567 & 0.9449 & 2.7085 & 1.6267\\
        d & 0.9440 & 0.9854 & 0.9369 & 0.7995 & 0.7276\\
        \bottomrule
    \end{tabular}
    \vspace{-5mm}
\end{table}

\subsection{Human Interaction}
\begin{figure*}[t!]
    \centering
    % \vspace{-0.6cm}
    \includegraphics[width= 1.8\columnwidth, trim=0.0cm 0cm 0cm 0cm, clip ]{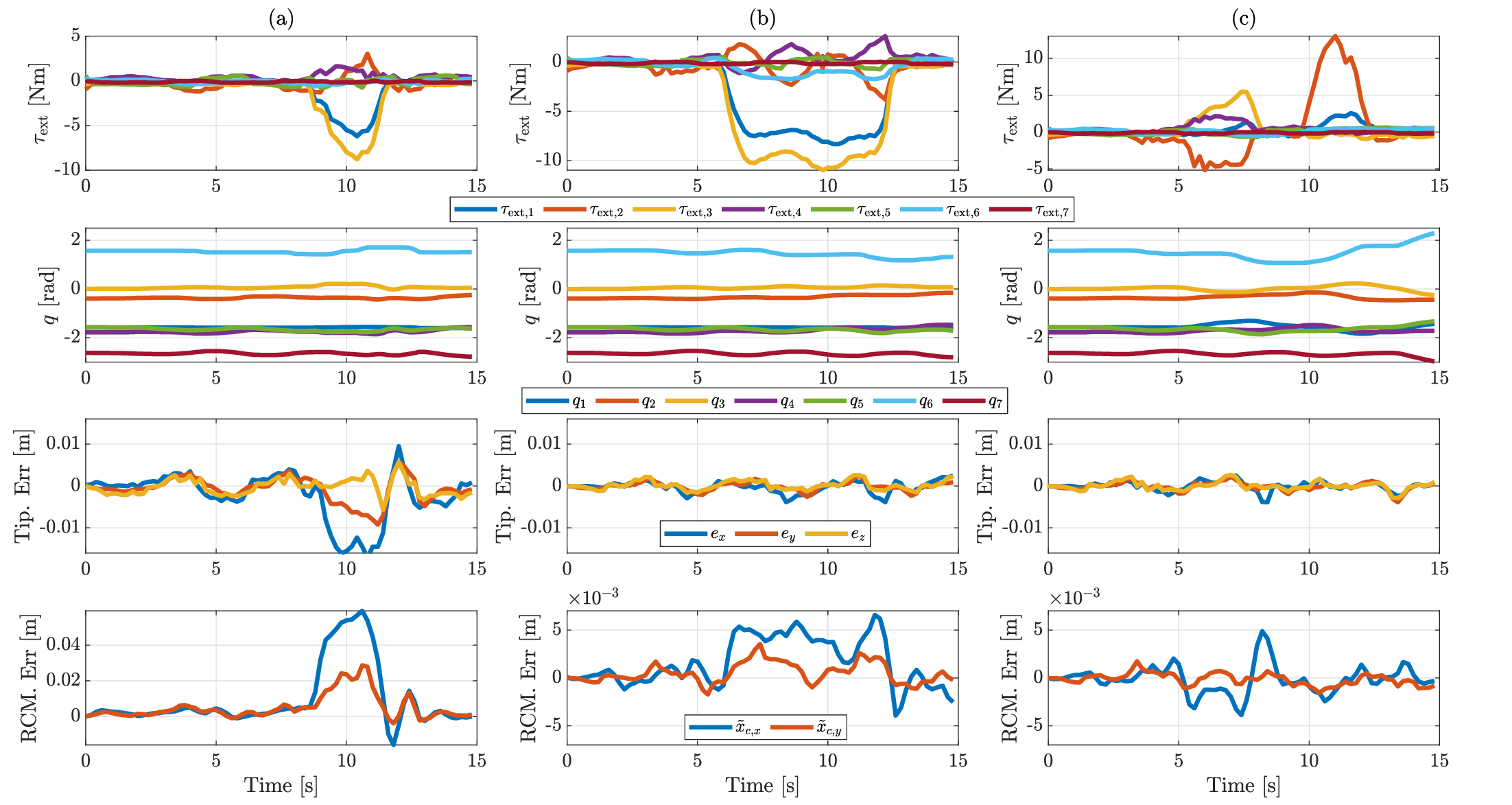}
    % \vspace{-0.4cm}
    \caption{Experimental interaction scenarios. From top to bottom, the panels show joint torques and joint positions, tool-tip tracking error, and projected RCM residual. (a) External interaction at the flange without disturbance compensation. (b) External interaction at the flange with disturbance compensation. (c) Intentional human interaction expressed through null-space compliance.}
    \label{fig:human-inter}
    % \vspace{-0.0 mm}
\end{figure*}

As shown by Sadeghian et al.~\cite{sadeghian2013dynamic,sadeghian2013task}, compensating external torques enhances disturbance rejection in both primary and secondary tasks. We conducted experiments in which the manipulator followed the same spiral trajectory of 20\,s duration at 50\% insertion depth. Incorporating the observed external torques increased robustness, as demonstrated in Fig.~\ref{fig:human-inter}(b), compared to Fig.~\ref{fig:human-inter}(a) without $\hat{\boldsymbol{\tau}}_{\mathrm{ext}}$. When an external torque was exerted at the flange, both tracking and RCM errors increased slightly but remained within acceptable limits.  

Intentional interaction was not suppressed but expressed through null-space compliance according to~\eqref{eqn:tau-nullimp}. In the experiment, push--pull forces were applied to the second link of the robot, mimicking arm bending into different attack angles (see Fig.~\ref{fig:robot-frame}). The observed torque at joint~2 and the corresponding induced torque at joint~1 are shown in Fig.~\ref{fig:human-inter}(c). A joint stiffness of $5\,\mathrm{Nm/rad}$ with sufficient damping was applied. Several joints, notably $q_1$, $q_3$, $q_5$, and $q_6$, actively contributed to the compliant null-space response at different phases of motion. In theory, null-space compliance should not degrade tip tracking or RCM satisfaction; in practice, however, a small increase in errors is observed in Fig.~\ref{fig:human-inter}(c) compared to the baseline in Fig.~\ref{fig:human-inter}(a).

\section{Discussion and conclusion} \label{sec:conclusion}

\textit{Discussion.}  
This work formulates RCM enforcement as a rheonomic holonomic constraint within a projected constraint-consistent controller. Across the tested simulation and hardware scenarios, the proposed method provides lower RCM residuals and smoother torque profiles than the considered baselines, while retaining accurate tool-tip tracking and enabling null-space compliance. These improvements arise from the explicit separation of constrained and free-motion actions in the control law. Several limitations, however, remain.

First, projection-based formulations remain numerically sensitive~\cite{moura2019equivalence}. Even small modeling errors or ill-conditioned Jacobians can destabilize the projector. Second, RCM kinematics are inherently depth-sensitive: due to the lever-arm effect, control gains tuned for one insertion length may degrade performance at another, making it difficult to define a unified gain set. Third, although a PD(+) law was used for fair comparison, passivity-preserving extensions remain an interesting direction for future work. Fourth, while torque smoothness has improved, residual high-frequency oscillations may degrade accuracy with long or flexible instruments in practical deployment, and the lack of direct tip sensing means forward kinematics remain the only feedback source. Fifth, null-space compliance, though enabling interaction, is configuration-dependent: in some poses, surgeons may not intuitively adjust the attack angle, suggesting that admittance control is still relevant. Finally, gain tuning itself poses a practical challenge: despite theoretical orthogonality, RCM and tip-task gains remain coupled, and adjusting one often affects the other. This interdependence was particularly evident in the \(\boldsymbol{Z}\)- and U--K baselines and complicates fair cross-method comparison.

\textit{Conclusion.}  
The proposed controller provides a constraint-consistent torque-control formulation for RCM-constrained surgical robotics and demonstrates improved joint-torque smoothness, reduced torque demand, and lower RCM residuals in the tested scenarios. Future research should focus on reducing numerical sensitivity, mitigating depth dependence, integrating passivity-preserving designs, and exploring shared-control schemes that accommodate inequivalent task constraints.

\bibliography{review}  % .bib

@article{smith2005robotic,
  title={Robotic-assisted laparoscopic prostatectomy: do minimally invasive approaches offer significant advantages?},
  author={Smith Jr, Joseph A and Herrell, S Duke},
  journal={Journal of clinical oncology},
  volume={23},
  number={32},
  pages={8170--8175},
  year={2005},
  publisher={American Society of Clinical Oncology}
}

@phdthesis{nugent2012evaluation,
  title={The evaluation of fundamental ability in acquiring minimally invasive surgical skill sets},
  author={Nugent, Emmeline},
  year={2012},
  school={Royal College of Surgeons in Ireland}
}

@article{elek2019robot,
  title={Robot-assisted minimally invasive surgical skill assessment—Manual and automated platforms},
  author={Elek, R Nagyn{\'e} and Haidegger, Tam{\'a}s},
  journal={Acta Polytechnica Hungarica},
  volume={16},
  number={8},
  pages={141--169},
  year={2019}
}

@inproceedings{moura2019equivalence,
  title={Equivalence of the projected forward dynamics and the dynamically consistent inverse solution},
  author={Moura, Joao and Ivan, Vladimir and Erden, Mustapha Suphi and Vijayakumar, Sethu},
  booktitle={Robotics: Science and Systems 2019},
  pages={1--10},
  year={2019}
}

@article{wang2022family,
  title={A family of RCM mechanisms: type synthesis and kinematics analysis},
  author={Wang, Zhi and Zhang, Wuxiang and Ding, Xilun},
  journal={International Journal of Mechanical Sciences},
  volume={231},
  pages={107590},
  year={2022},
  publisher={Elsevier}
}

@article{aksungur2015remote,
  title={Remote center of motion (RCM) mechanisms for surgical operations},
  author={Aksungur, Serhat},
  journal={International Journal of Applied Mathematics Electronics and Computers},
  volume={3},
  number={2},
  pages={119--126},
  year={2015},
  publisher={{\.I}smail SARITA{\c{S}}}
}

@article{funda1996constrained,
  title={Constrained Cartesian motion control for teleoperated surgical robots},
  author={Funda, Janez and Taylor, Russell H and Eldridge, Ben and Gomory, Stephen and Gruben, Kreg G},
  journal={IEEE Transactions on Robotics and Automation},
  volume={12},
  number={3},
  pages={453--465},
  year={1996},
  publisher={IEEE}
}

@inproceedings{aghakhani2013task,
  title={Task control with remote center of motion constraint for minimally invasive robotic surgery},
  author={Aghakhani, Nastaran and Geravand, Milad and Shahriari, Navid and Vendittelli, Marilena and Oriolo, Giuseppe},
  booktitle={2013 IEEE international conference on robotics and automation},
  pages={5807--5812},
  year={2013},
  organization={IEEE}
}

@inproceedings{kastritsi2021control,
  title={A control method for time-variant RCM constraint in hands-on ramis procedures},
  author={Kastritsi, Theodora and Doulgeri, Zoe},
  booktitle={2021 29th Mediterranean Conference on Control and Automation (MED)},
  pages={729--734},
  year={2021},
  organization={IEEE}
}

@article{sadeghian2019constrained,
  title={Constrained kinematic control in minimally invasive robotic surgery subject to remote center of motion constraint},
  author={Sadeghian, Hamid and Zokaei, Fatemeh and Hadian Jazi, Shahram},
  journal={Journal of intelligent \& robotic systems},
  volume={95},
  pages={901--913},
  year={2019},
  publisher={Springer}
}

@article{davila2024real,
  title={Real-time inverse kinematics for robotic manipulation under remote center-of-motion constraint using memetic evolution},
  author={Davila, Ana and Colan, Jacinto and Hasegawa, Yasuhisa},
  journal={Journal of Computational Design and Engineering},
  volume={11},
  number={3},
  pages={248--264},
  year={2024},
  publisher={Oxford University Press}
}

@article{piccinelli2024passive,
  title={A passive convex optimal control algorithm for teleoperating a redundant robotic arm in minimally invasive surgery},
  author={Piccinelli, Nicola and Colombo-Taccani, Gianluca and Muradore, Riccardo},
  journal={International Journal of Robust and Nonlinear Control},
  year={2024},
  publisher={Wiley Online Library}
}

@article{kastritsi2024passive,
  title={Passive Bilateral Surgical Teleoperation With RCM and Spatial Constraints in the Presence of Time Delays},
  author={Kastritsi, Theodora and Semetzidis, Theofanis Prapavesis and Doulgeri, Zoe},
  journal={IEEE Transactions on Robotics},
  year={2024},
  publisher={IEEE}
}

@inproceedings{sandoval2017new,
  title={A new kinematic formulation of the RCM constraint for redundant torque-controlled robots},
  author={Sandoval, Juan and Poisson, G{\'e}rard and Vieyres, Pierre},
  booktitle={2017 IEEE/RSJ International Conference on Intelligent Robots and Systems (IROS)},
  pages={4576--4581},
  year={2017},
  organization={IEEE}
}

@article{su2019improved,
  title={Improved human--robot collaborative control of redundant robot for teleoperated minimally invasive surgery},
  author={Su, Hang and Yang, Chenguang and Ferrigno, Giancarlo and De Momi, Elena},
  journal={IEEE Robotics and Automation Letters},
  volume={4},
  number={2},
  pages={1447--1453},
  year={2019},
  publisher={IEEE}
}

@article{su2022fuzzy,
  title={Fuzzy approximation-based task-space control of robot manipulators with remote center of motion constraint},
  author={Su, Hang and Qi, Wen and Chen, Jiahao and Zhang, Dandan},
  journal={IEEE Transactions on Fuzzy Systems},
  volume={30},
  number={6},
  pages={1564--1573},
  year={2022},
  publisher={IEEE}
}

@article{li2020accelerated,
  title={An accelerated finite-time convergent neural network for visual servoing of a flexible surgical endoscope with physical and RCM constraints},
  author={Li, Weibing and Chiu, Philip Wai Yan and Li, Zheng},
  journal={IEEE transactions on neural networks and learning systems},
  volume={31},
  number={12},
  pages={5272--5284},
  year={2020},
  publisher={IEEE}
}

@article{liu2024data,
  author={Liu, Mei and Liu, Kun and Zhu, Puchen and Zhang, Guoqian and Ma, Xin and Shang, Mingsheng},
  journal={IEEE Transactions on Industrial Informatics}, 
  title={{Data-Driven Remote Center of Cyclic Motion (RC$^{2}$M) Control for Redundant Robots With Rod-Shaped End-Effector}}, 
  year={2024},
  volume={20},
  number={4},
  pages={6772-6780},
  doi={10.1109/TII.2024.3353930}}

@article{aghili2005unified,
  title={A unified approach for inverse and direct dynamics of constrained multibody systems based on linear projection operator: applications to control and simulation},
  author={Aghili, Farhad},
  journal={IEEE Transactions on Robotics},
  volume={21},
  number={5},
  pages={834--849},
  year={2005},
  publisher={IEEE}
}

@inproceedings{aghili2016control,
  title={Control of constrained robots subject to unilateral contacts and friction cone constraints},
  author={Aghili, Farhad and Su, Chun-Yi},
  booktitle={2016 IEEE international conference on robotics and automation (ICRA)},
  pages={2347--2352},
  year={2016},
  organization={IEEE}
}

@inproceedings{wang2023towards,
  title={Towards exact interaction force control for underactuated quadrupedal systems with orthogonal projection and quadratic programming},
  author={Wang, Shengzhi and Chu, Xiangyu and Au, KW Samuel},
  booktitle={2023 IEEE International Conference on Robotics and Automation (ICRA)},
  pages={12268--12274},
  year={2023},
  organization={IEEE}
}

@inproceedings{mistry2012operational,
  title={Operational space control of constrained and underactuated systems},
  author={Mistry, Michael and Righetti, Ludovic},
  booktitle={Robotics: Science and systems},
  volume={7},
  pages={225--232},
  year={2012}
}

@inproceedings{righetti2011inverse,
  title={Inverse dynamics control of floating-base robots with external constraints: A unified view},
  author={Righetti, Ludovic and Buchli, Jonas and Mistry, Michael and Schaal, Stefan},
  booktitle={2011 IEEE international conference on robotics and automation},
  pages={1085--1090},
  year={2011},
  organization={IEEE}
}

@article{udwadia2002foundations,
  title={On the foundations of analytical dynamics},
  author={Udwadia, Firdaus E and Kalaba, Robert E},
  journal={International Journal of non-linear mechanics},
  volume={37},
  number={6},
  pages={1079--1090},
  year={2002},
  publisher={Elsevier}
}

@inproceedings{minelli2021dynamic,
  title={Dynamic-based RCM torque controller for robotic-assisted minimally invasive surgery},
  author={Minelli, Marco and Secchi, Cristian},
  booktitle={2021 IEEE/RSJ International Conference on Intelligent Robots and Systems (IROS)},
  pages={733--740},
  year={2021},
  organization={IEEE}
}

@inproceedings{minelli2022torque,
  title={A torque controlled approach for virtual remote centre of motion implementation},
  author={Minelli, Marco and Secchi, Cristian},
  booktitle={2022 IEEE/RSJ International Conference on Intelligent Robots and Systems (IROS)},
  pages={4949--4956},
  year={2022},
  organization={IEEE}
}

@article{khatib1987unified,
  title={A unified approach for motion and force control of robot manipulators: The operational space formulation},
  author={Khatib, Oussama},
  journal={IEEE Journal on Robotics and Automation},
  volume={3},
  number={1},
  pages={43--53},
  year={1987},
  publisher={IEEE}
}

@inproceedings{kenanoglu2024design,
  title={Design and Evaluation of a Surgical Tool Drive Unit for Sustainable Training in Robot-Assisted Minimally Invasive Surgery},
  author={Kenanoglu, Celal Umut and Le Mesle, Valentin and Luarasi, Gjergji and Sadeghian, Hamid and Haddadin, Sami},
  booktitle={2024 10th IEEE RAS/EMBS International Conference for Biomedical Robotics and Biomechatronics (BioRob)},
  pages={1555--1560},
  year={2024},
  organization={IEEE}
}

@article{sadeghian2013dynamic,
  title={Dynamic multi-priority control in redundant robotic systems1},
  author={Sadeghian, Hamid and Villani, Luigi and Keshmiri, Mehdi and Siciliano, Bruno},
  journal={Robotica},
  volume={31},
  number={7},
  pages={1155--1167},
  year={2013},
  publisher={Cambridge University Press}
}

@inproceedings{de2006collision,
  title={Collision detection and safe reaction with the DLR-III lightweight manipulator arm},
  author={De Luca, Alessandro and Albu-Schaffer, Alin and Haddadin, Sami and Hirzinger, Gerd},
  booktitle={2006 IEEE/RSJ international conference on intelligent robots and systems},
  pages={1623--1630},
  year={2006},
  organization={IEEE}
}

@book{ott2008cartesian,
  title={Cartesian impedance control of redundant and flexible-joint robots},
  author={Ott, Christian},
  year={2008},
  publisher={Springer}
}

@article{sadeghian2013task,
  title={Task-space control of robot manipulators with null-space compliance},
  author={Sadeghian, Hamid and Villani, Luigi and Keshmiri, Mehdi and Siciliano, Bruno},
  journal={IEEE Transactions on Robotics},
  volume={30},
  number={2},
  pages={493--506},
  year={2013},
  publisher={IEEE}
}

@article{dietrich2015passivation,
  title={Passivation of projection-based null space compliance control via energy tanks},
  author={Dietrich, Alexander and Ott, Christian and Stramigioli, Stefano},
  journal={IEEE Robotics and automation letters},
  volume={1},
  number={1},
  pages={184--191},
  year={2015},
  publisher={IEEE}
}

@article{kirilova2009visco,
  title={Visco-elastic mechanical properties of human abdominal fascia},
  author={Kirilova, Miglena and Stoytchev, Stoyan and Pashkouleva, Dessislava and Tsenova, Vesselina and Hristoskova, Radka},
  journal={Journal of bodywork and movement therapies},
  volume={13},
  number={4},
  pages={336},
  year={2009}
}

\end{document}